\documentclass[sn-mathphys-ay]{sn-jnl}

\usepackage{graphicx}%
\usepackage{multirow}%
\usepackage{amsmath,amssymb,amsfonts}%
\usepackage{amsthm}%
\usepackage{mathrsfs}%
\usepackage[title]{appendix}%
\usepackage{xcolor}%
\usepackage{textcomp}%
\usepackage{manyfoot}%
\usepackage{booktabs}%
\usepackage{algorithm}%
\usepackage{algorithmicx}%
\usepackage{algpseudocode}%
\usepackage{listings}%
\usepackage{longtable}
\usepackage{float}
\usepackage{hyperref} 
\usepackage{array}

\theoremstyle{thmstyleone}%
\theoremstyle{thmstyletwo}%

\theoremstyle{thmstylethree}%

\begin{document}

\title[Machine Learning Innovations in CPR: A Comprehensive Survey on Enhanced Resuscitation Techniques]{Machine Learning Innovations in CPR: A Comprehensive Survey on Enhanced Resuscitation Techniques}


\author[1]{\fnm{Saidul} \sur{Islam}}\email{saidul.islam@concordia.ca}

\author[2,1]{\fnm{Gaith} \sur{Rjoub}}\email{grjoub@aut.edu.jo}

\author[1]{\fnm{Hanae} \sur{Elmekki}}\email{hanae.elmekki@mail.concordia.ca}

\author*[3,1]{\fnm{Jamal} \sur{Bentahar}}\email{jamal.bentahar@concordia.ca}

\author[4,5,6]{\fnm{Witold} \sur{Pedrycz}}\email{wpedrycz@ualberta.ca}

\author[7]{\fnm{Robin} \sur{Cohen}}\email{rcohen@uwaterloo.ca}

\affil*[1]{\orgdiv{Concordia Institute for Information Systems Engineering}, \orgname{Concordia University}, \city{Montreal}, \country{Canada}}

\affil[2]{\orgdiv{Faculty of Information Technology}, \orgname{Aqaba University of Technology}, \city{Aqaba}, \country{Jordan}}

\affil[3]{\orgdiv{Department of Computer Science}, \orgname{Khalifa University}, \city{Abu Dhabi},  \country{UAE}}

\affil[4]{\orgdiv{Department of Electrical and Computer Engineering}, \orgname{University of Alberta}, \city{Edmonton},  \country{Canada}}

\affil[5]{\orgdiv{Systems Research Institute}, \orgname{Polish Academy of Sciences}, \city{Warsaw},  \country{Poland}}

\affil[6]{\orgdiv{Research Center of Performance and Productivity Analysis}, \orgname{Istinye University}, \city{Sariyer/Istanbul},  \country{Turkey}}

\affil[7]{\orgdiv{David R. Cheriton School of Computer Science}, \orgname{University of Waterloo}, \city{Waterloo},  \country{Canada}}

\abstract{This survey paper explores the transformative role of Machine Learning (ML) and Artificial Intelligence (AI) in Cardiopulmonary Resuscitation (CPR). It examines the evolution from traditional CPR methods to innovative ML-driven approaches, highlighting the impact of predictive modeling, AI-enhanced devices, and real-time data analysis in improving resuscitation outcomes. The paper provides a comprehensive overview, classification and critical analysis of current applications, challenges, and future directions in this emerging field.
}




\maketitle

\section{Introduction}
\label{sec1}

Cardiopulmonary Resuscitation (CPR) is a life-saving medical procedure that has been instrumental in the field of emergency medicine for several decades~\citep{hurt2005modern}. Originating as a technique to revive individuals from drowning incidents, it has evolved into a universally recognized procedure for cardiac arrest victims, regardless of the cause~\citep{cooper2006cardiopulmonary}. At its core, CPR serves as an interim measure to simulate the heart's function of pumping blood, ensuring that oxygenated blood continues to circulate to vital organs, especially the brain~\citep{raza2021}. This is crucial because brain cells begin to die within minutes without oxygen, leading to irreversible brain damage or death~\citep{lee2022}. With more than 400,000 cases each year in North America and an average survival rate of $10\%$, the impact of Out-of-Hospital Cardiac Arrest (OHCA) on public health is substantial \citep{meier2010chest}. Despite recent technological advancements, the survival rate has been stagnating over the past few years. Management of cardiac arrest is defined by international recommendations, defined by National Resuscitation Council \citep{herlitz1997rhythm}.  It is optimized by multiple interconnected links that form the Chain of Survival.  One of the most important link of this chain is CPR \citep{chamberlain2008chest}. The American Heart Association and other global health organizations have emphasized the significance of CPR, not just among healthcare professionals but also among the general public~\citep{hinkelbein2020}. This is because the majority of cardiac arrests occur outside of hospital settings, where immediate medical intervention is not readily available. In such scenarios, bystander CPR can be the difference between life and death~\citep{pellegrino2021}. Despite its proven efficacy, several challenges persist in the realm of CPR. These include disparities in public awareness, variations in training methodologies, and evolving guidelines based on new research findings~\citep{raza2021}.

While traditional CPR techniques undoubtedly save countless lives and play a crucial role in the field of medical emergencies, several challenges persist. Specifically, the effectiveness of traditional CPR is hindered by the variability in human performance. Due to the dynamic nature of CPR in medical emergencies, it is very difficult for humans to consistently provide optimal pressure at the appropriate time intervals over an extended period~\citep{challange00}. The recent advancements in technology and the emergence of Machine Learning (ML) offer new possibilities. To address these challenges, ML presents a potential solution to assist medical practitioners and enhance resuscitation techniques. ML algorithms can process vast amounts of data and make real-time decisions, potentially improving the accuracy and efficiency of CPR interventions. Integrating ML into CPR procedures could revolutionize the way we respond to cardiac arrests and significantly improve patient outcomes ~\citep{dahal2022overview, challange01}.

The paper methodically reviews and analyzes existing ML applications in CPR, identifies research gaps, and uncovers unexplored potential, setting a new direction for empirical studies and technological advancements. This comprehensive synthesis is designed to spark innovative research and collaborations, with the goal of significantly advancing CPR techniques, ultimately impacting clinical outcomes in emergency medicine. Having noticed a gap in the existing literature that surveys exploring the integration of  ML in CPR, our work aims to provide a comprehensive
interdisciplinary reference. The use of CPR in emergency medicine faces several challenges due to variability in human performance, with the need to respond under pressure to problems that are dynamically changing. This paper establishes a framework for future research that demonstrates the value of ML in assisting medical practitioners and improving resuscitation techniques. The unexplored potential we uncover sets new directions for empirical studies and technological advancements. By providing this framework, we aim to contribute a valuable and novel perspective to the field.

\begin{figure}[!htp]
\centering
\includegraphics[width=\textwidth]{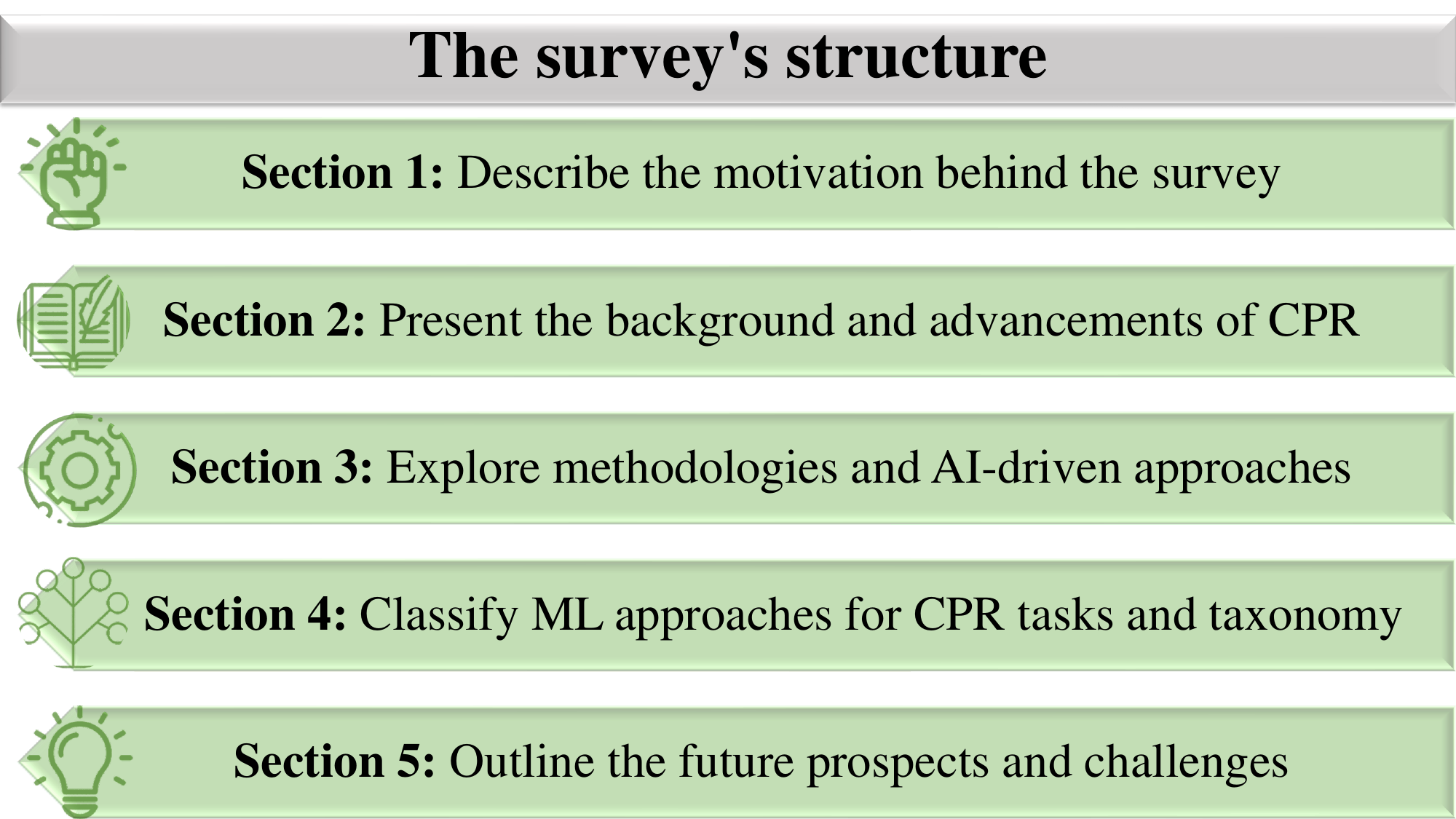}
\caption{Visual Abstract}
\label{visual_abstract}
\end{figure}

The objective of this survey is to highlight the value of applying ML in this field of CPR, set against the historical background of the method, its current practices, and its potential to support new techniques. The paper is structured as follows: Section \ref{section2} provides the historical background of CPR. We present an overview of the techniques and the advancement of CPR in Section \ref{section3}. In Section \ref{section4}, we propose the classification of CPR studies available in the literature. Finally, in Section \ref{section5}, we conclude the paper by summarizing our findings and discussing potential avenues for future research. Figure \ref{visual_abstract} gives a visual abstract of the whole organization of the proposed paper. The detailed abbreviations and definitions used in the paper are listed in Table \ref{tab:abbr}

\begin{table} [!htp]
  \centering
  \small
  \begin{tabular}{| >{\centering\arraybackslash}p{3cm} | >{\centering\arraybackslash}p{10cm}|}
    \hline
    \textbf{Abbreviation} & \textbf{Definition} \\ \hline
    AED & Automated External Defibrillator \\ \hline
    AI & Artificial Intelligence \\ \hline
    ANN & Artificial Neural Network \\ \hline 
    CNN & Convolutional Neural Network \\ \hline
    CPR & Cardiopulmonary Resuscitation \\ \hline
    ECG & Electrocardiogram \\ \hline
    ECPR & Extracorporeal CPR \\ \hline
    KNN & K-Nearest Neighbors \\ \hline
    LSTM & Long Short-Term Memory \\ \hline
    ML & Machine Learning \\ \hline
    OHCA & Out-of-Hospital Cardiac Arrest \\ \hline
    POCUS & Point-of-Care Ultrasound \\ \hline
    REBOA & Resuscitative Endovascular Balloon Occlusion of the Aorta \\ \hline
    RQI & Resuscitation Quality Improvement \\ \hline
    RL & Reinforcement Learning \\ \hline
    ROSC & Return Of Spontaneous Circulation Detection \\ \hline
    SVM & Support Vector Machine \\ \hline
    XAI & Explainable Artificial Intelligence \\ \hline
  \end{tabular}
  \caption{List of abbreviations and acronyms used in the paper}
  \label{tab:abbr}
\end{table}

\section{Background}

\subsection{Historical Background}
\label{section2}
\begin{figure}[!htp]
\centering
\includegraphics[width=\textwidth]{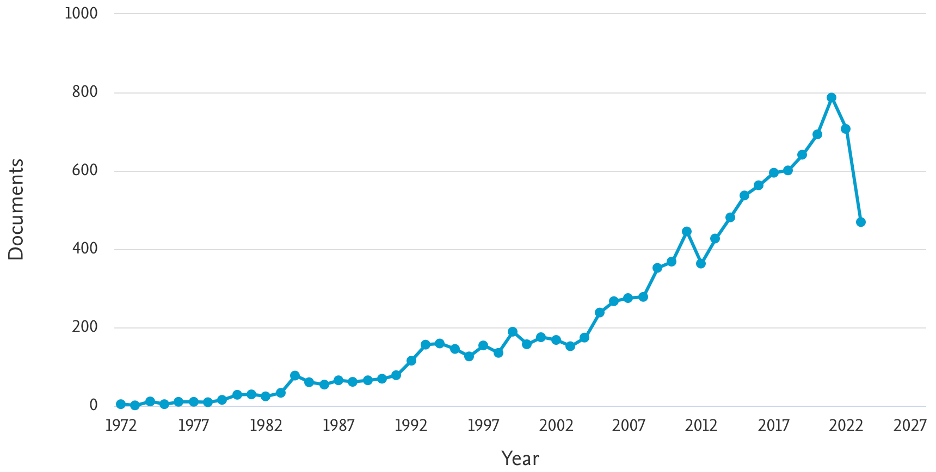}
\caption{Evolution of published CPR documents over the years.}
\label{fig1}
\end{figure}

The history of CPR has evolved dramatically from early airway management techniques in the 16th century to systematic approaches developed in the 18th century. Key advancements include the refinement of mouth-to-mouth resuscitation in the late 1950s and the progression of cardiac massage from open chest methods in 1874 to closed chest techniques by 1960. Significant improvements in electrical defibrillation also occurred, starting with internal use in 1947 and moving to external applications by 1956 ~\citep{cooper2006cardiopulmonary, debard1980history}. These modern CPR methods were standardized in the past few decades, leading to the development of automatic and semi-automatic defibrillators, which have enhanced prehospital and home resuscitation efforts. The integration of these techniques has significantly improved survival rates from cardiac arrest, illustrating the critical importance of ongoing research and innovation in CPR ~\citep{cummins1986automatic}. Today, CPR has become a globally recognized emergency procedure, continuously refined through research and technological advancements. Recent innovations in ML and AI analyze real-time data to enhance the precision of CPR, improving outcomes. These advancements promise to further increase CPR's effectiveness and save more lives. Figure \ref{fig1} is a time-series line chart showing the number of CPR-related research publications over the years. The graph indicates increasing interest in CPR research, with plateaus suggesting periods of methodological refinement.

\begin{figure}[!htp]
\centering
\includegraphics[width=\textwidth]{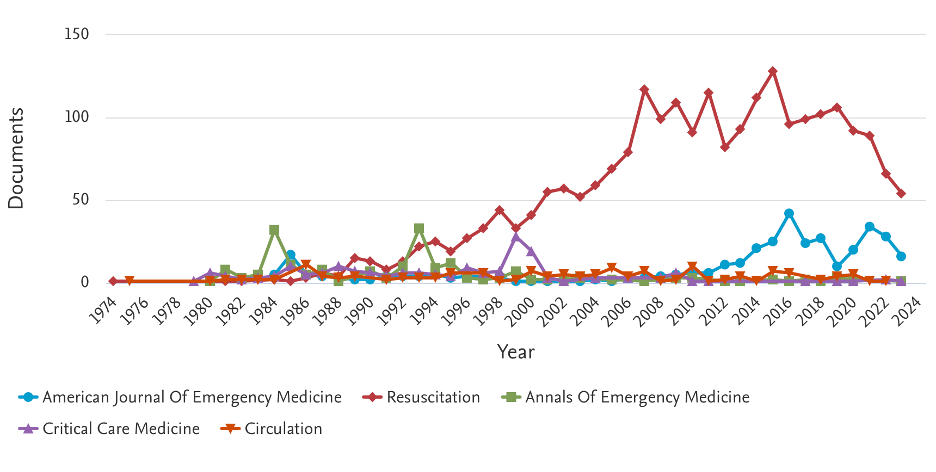}
\caption{Number of CPR-related documents over the years from various sources.}
\label{fig2}
\end{figure}

\begin{figure}[!htp]
\centering
\includegraphics[width=\textwidth]{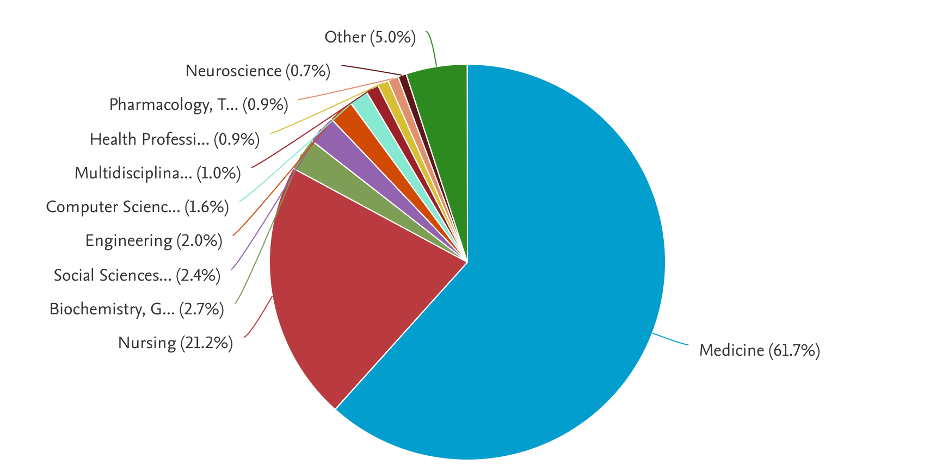}
\caption{Document contributions from various subject areas to CPR research}
\label{fig3}
\end{figure}

Moreover, Figure \ref{fig2} shows the number of CPR-related documents over the years with each line representing a source. The rise of a line signifies that the source had more interest in CPR that year, when the line goes down  mean the source was refining their work or changing their research focus. In addition, this figures show how CPR research changes over time, in some years there is a shared focus in the reserach community, whereas for other years the lines are spread out which could mean different approaches or new areas of exploration by various sources. Furthermore, Figure \ref{fig3} presents the subject areas and their contributions to CPR knowledge. Based on the figure, areas like cardiology or emergency medicine have more research publications in CPR. The figure depicts also the interdisciplinary nature of the CPR field, with several domains intervening from technological and medical perspectives.

\subsection{Historical Techniques}

Over time, CPR has evolved alongside advancements in medicine. By the end of the $19$th and early $20$th centuries, the role of oxygenation and circulation became increasingly important to the success of resuscitation efforts. However, the most significant evolution in CPR occurred with the introduction of closed-chest cardiac massage combined with mouth-to-mouth resuscitation.

\begin{itemize}
    
    \item \textbf{Mouth-to-mouth resuscitation:} Mouth-to-mouth resuscitation, widely recognized as a lifesaving technique in the $20$th century, actually has roots that trace back to ancient practices. Studies, such as the one conducted by~\cite{poulsen1959pulmonary}, have demonstrated that this method can provide adequate alveolar ventilation. However, it also carries risks, including the possibility of hypoventilation. This highlights the importance of proper training in different resuscitation techniques to ensure effective emergency responses.

     \item \textbf{Closed-Chest Cardiac Massage:} In the mid-$20$th century, \cite{kouwenhoven1960} demonstrated that rhythmic chest compression could generate external blood flow. This non-invasive method, replacing manual heart massage through surgery, allows for easy, equipment-free performance, significantly improving OHCA survival rates. Despite potential cardiac injuries~\citep{baldwin1976,park2017}, the benefits outweigh the risks compared to no intervention.

\end{itemize}

\subsection{Modern Advancements}

The landscape of CPR has witnessed profound transformations in recent decades, with the advancements in medical science leading to improved techniques and tools for first responders and medical professionals, expanding life-saving possibilities.

\begin{itemize}
    \item \textbf{Automated External Defibrillators (AEDs):} All studied AEDs meet diagnostic rhythm performance standards in real OHCA scenarios, with errors typically due to operator issues rather than device faults~\citep{zijlstra2017automated}. Fully automatic AEDs enhance correct shock delivery, and improving emergency cardiac care outcomes.

    \item \textbf{Compression-Only CPR:} Compression-only CPR maintains organ blood flow during cardiac arrest and simplifies the process by eliminating mouth-to-mouth ventilation. Studies show compressions are crucial in early cardiac arrest for sufficient blood oxygen~\citep{sayre2008}.

    \item \textbf{Mechanical CPR Devices:} Mechanical CPR devices, which deliver consistent and optimal compressions, reduce the physical exertion required from rescuers. They are especially beneficial during patient transport or challenging manual compression situations~\citep{rubertsson2014}.

    \item \textbf{Point-of-Care Ultrasound (POCUS):} POCUS has transformed cardiac arrest management by enabling real-time heart visualization, allowing clinicians to quickly identify and treat reversible causes. It also guides interventions like fluid administration and medication adjustments based on real-time cardiac responses~\citep{gaspari2016}.

    \item \textbf{Feedback Devices in CPR:} \cite{skorning2010new} describe a visual feedback device that improves chest compression performance in CPR simulations, enhancing rate, depth, and sufficiency. Its simplicity makes it suitable for trained responders, reinforcing the chain of survival and potentially improving cardiac arrest outcomes.

    \item \textbf{Telemedicine and Remote Guidance:} Telemedicine integrates into CPR by providing real-time guidance via video calls, notably improving chest compression rate and accuracy over audio-only instructions. However, it may slightly delay bystander CPR initiation in simulated settings~\citep{lin2018quality}.

    \item \textbf{Integration of AR-VR Technology:} AR and VR technologies expand CPR training by enabling practice in realistic scenarios, yet wider adoption is needed in this promising research area~\citep{semsarian2016}.

    \end{itemize}

\section{CPR Methodologies and AI-Driven Approaches}
\label{section3}

\subsection{Methodologies}

The evolving field of CPR is marked by numerous impactful studies across different years, employing varied methodologies and focusing on diverse applications. A comparative analysis is necessary to explore the insights into CPR technique evolution and current practices, highlighting the ongoing importance of research to enhance and optimize CPR procedures. Table \ref{table1} provides a comprehensive comparison of several studies that have contributed to the field of CPR. Each study is categorized based on its publication year, the methodology employed, its primary application, and the impact it has had on the field. Based on this table, we gain an idea about the evolution of CPR techniques and the current state of CPR, with a highlight of the diverse methodologies and applications used. The table also shows the importance of continuing research to enhance the field of CPR.

\setlength\LTleft{-1.7cm}
\begin{longtable}
{|>{\centering\arraybackslash}p{2.2cm}|>{\centering\arraybackslash}p{2cm}|>{\centering\arraybackslash}p{3cm}|>{\centering\arraybackslash}p{3cm}|>{\centering\arraybackslash}p{4.5cm}|> {\centering\arraybackslash}p{1.5cm}| 
}

\caption{Detailed Comparison of References for CPR Advancements and Challenges.}\\
\hline 
\label{table1}
\textbf{Reference} & \textbf{Year} & \textbf{Methodology} & \textbf{Application} & \textbf{Impact} \\
\hline
\endfirsthead

\multicolumn{6}{c}%
{{\bfseries \tablename\ \thetable{} -- continued from previous page}} \\
\hline
\textbf{Reference} & \textbf{Year} & \textbf{Methodology} & \textbf{Application} & \textbf{Impact} \\
\hline
\endhead

\hline \multicolumn{6}{|r|}{{Continued on next page}} \\
\hline
\endfoot

\hline
\endlastfoot

\cite{zijlstra2017automated} & 1998 & Observational study & Public settings & Increased survival rates with AEDs \\
\hline
\cite{sayre2008} & 2008 & Literature review & Early-stage CPR & Compression-only CPR as effective as traditional \\
\hline
\cite{hunt2014} & 2014 & Meta-analysis & CPR training & Enhanced outcomes with high-fidelity manikins \\
\hline
\cite{teeter2018} & 2018 & Observational study & Trauma-induced cardiac arrest & Potential benefits of REBOA \\
\hline
\cite{white2018advanced} & 2018 & Meta-analysis & Advanced resuscitation & Improved outcomes with advanced airway management \\
\hline
\cite{rubertsson2014} & 2014 & Clinical trial & CPR delivery & Mechanical CPR devices as effective as manual \\
\hline
\cite{gaspari2016} & 2016 & Observational study & Cardiac arrest diagnosis & POCUS aids in determining arrest causes \\
\hline
\cite{nielsen2013} & 2013 & Clinical trial & Post-resuscitation care & Benefits of therapeutic hypothermia \\
\hline
\cite{bartos2020improved} & 2018 & Observational study & Refractory cardiac arrest & Benefits of ECPR \\
\hline
\cite{skorning2010new} & 2015 & Clinical trial & CPR training & Improved compression quality with feedback devices \\
\hline
\cite{blewer2017} & 2017 & Observational study & Bystander CPR & Increased CPR rates with public training \\
\hline
\cite{duff20182018} & 2018 & Literature review & Pediatric CPR & Unique considerations for pediatric patients \\
\hline
\cite{semsarian2016} & 2016 & Literature review & Sudden cardiac deaths & Technology aids in determining causes \\
\hline
\cite{thakur2023} & 2023 & Descriptive cross-sectional study & Neonatal Intensive Care Unit & Technological advancements improve neonatal survival \\
\hline
\cite{wyckoff2022} & 2022 & Literature review and consensus & CPR & Latest consensus on CPR and emergency care \\
\hline
\cite{aldridge2022} & 2022 & Scoping review & Bystander CPR during emergency calls & Barriers and facilitators to bystander CPR \\
\hline
\cite{ijuin2022} & 2022 & Observational study & ECPR in hybrid ER & Utility of hybrid ER for ECPR decisions \\
\hline
\end{longtable}

\subsection{AI-Driven Approaches}

\begin{figure}[!htp]
\centering
\includegraphics[width=\textwidth]{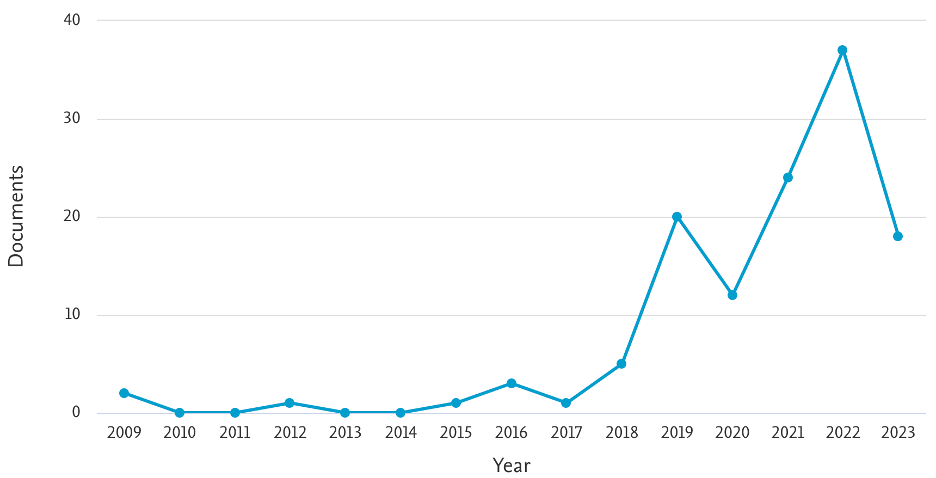}
\caption{Documents published per year showing the growing interest and adoption of AI and ML in CPR research}
\label{fig4}
\end{figure}

The application of ML has become increasingly prominent in medicine, as it supports healthcare professionals in analyzing and diagnosing various health conditions, with a particular focus on cardiovascular diseases \citep{bhushan2023machine, patel2022survey}. In particular, CPR is one area where ML is beginning to play a prominent role, especially with advancements in devices and technology. These developments are aiding in the refinement and automation of CPR techniques, leading to improvements in resuscitation practices. Figure \ref{fig4} illustrates the increasing adoption of AI and ML technologies in CPR research over recent years. The x-axis tracks the progression of years, while the y-axis quantifies the number of studies or documents, indicating a rising interest in leveraging these advanced algorithms for CPR enhancement. This trend reflects the benefits of AI and ML in data analysis, rhythm detection, and predictive modeling within resuscitation efforts, highlighting a promising frontier in the field.

\begin{figure}[!htp]
\centering
\includegraphics[width=\textwidth]{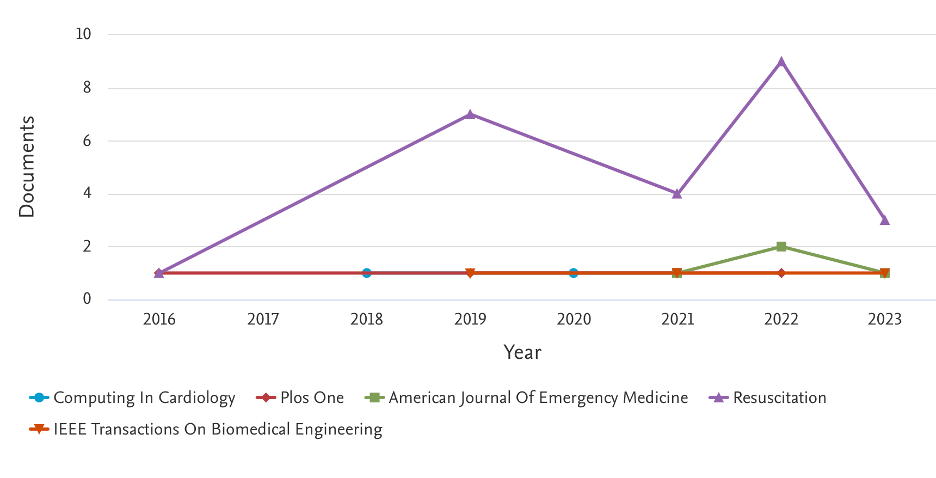}
\caption{Annual distribution of CPR-related documents using AI and ML, with each line representing different sources or publishers}
\label{fig5}
\end{figure}

In addition, Figure \ref{fig5} displays the yearly distribution of CPR-related documents utilizing AI and ML techniques from different sources or publishers. Each line in the plot corresponds to a specific source, highlighting periods of increased contribution to AI and ML-driven CPR research. The intertwining trajectories on this canvas not only narrate the contributions of individual sources but also highlight the collective evolution of the research community in embracing advanced computational techniques in CPR. In essence, Figure \ref{fig5} offers a panoramic view into the dynamic landscape of AI and ML in CPR research, emphasizing both individual and collaborative strides made by various sources over time.

\begin{figure}[!htp]
\centering
\includegraphics[width=\textwidth]{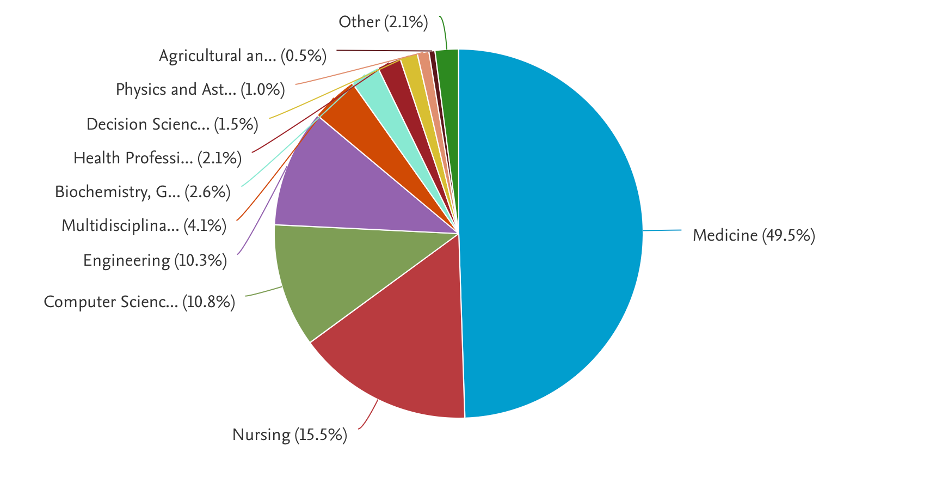}
\caption{Distribution of CPR-related documents using AI and ML across various subject areas}
\label{fig6}
\end{figure}

However, Figure \ref{fig6} presents the distribution of CPR-related documents that utilize AI and ML across varied subject areas. Each area is represented with a color, and the size represents the volume of research in that particular area. Thus, high volume represents areas that integrate AI and ML highly in CPR research, while the area with less volume may represent emerging fields where contributions are still evolving. These could include interdisciplinary areas where technology, data science, and medical research intersect. The collective ensemble of these segments paints a comprehensive picture of the diverse academic avenues exploring the confluence of AI, ML, and CPR. Figure \ref{fig6} accentuates the rich interdisciplinary fabric of modern CPR research, showcasing how varied domains contribute to the evolving narrative of resuscitation in the age of computation.

\section{Classification of ML Approaches for CPR Tasks}
\label{section4}

\subsection{Taxonomy}

\begin{figure} [!htp]
    \begin{center}
    \makebox[\textwidth][l]{\hspace*{-2.9cm}\includegraphics[width=1.45\textwidth,height=1.26\textwidth]{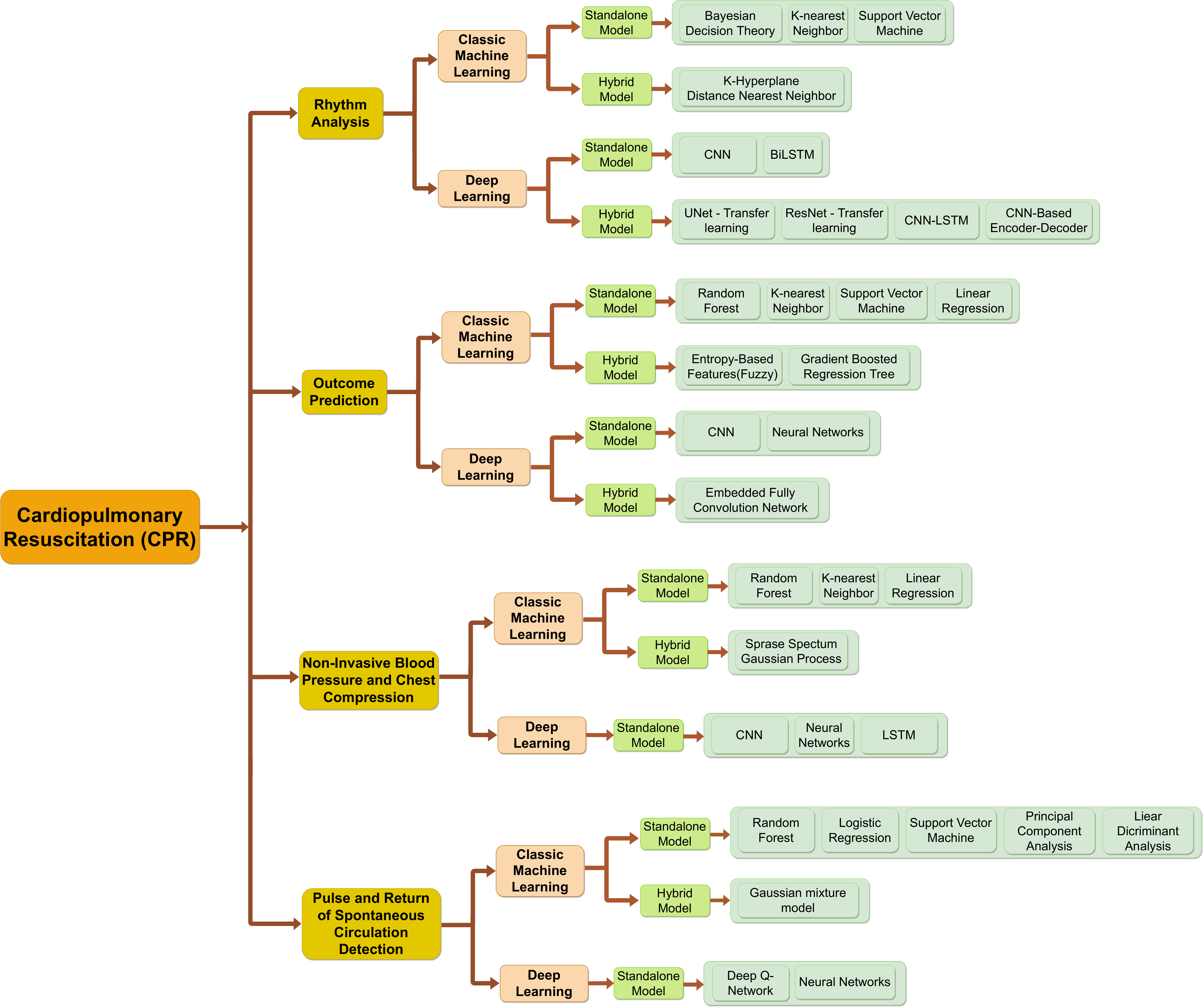}}
    \caption{Taxonomy of ML-driven CPR tasks.}
    \label{fig:taxonomy}
    \vspace{-10pt} 
    \end{center}
\end{figure}

The integration of ML and AI in CPR has led to various innovations. These advancements range from early detection of cardiac arrest during emergency calls to the optimization of chest compressions and ventilation. The upcoming part of this paper provides a comparative overview of several studies that have contributed to these machine-driven innovations in CPR. We present the ML techniques and classes covered in each publication along with the details of the application fields for each model, and their key contributions, and provide a concise summary of the paper. We conducted thorough research across different databases: Pubmed, Google Scholar, IEEE, Science Direct, and ACM to find papers introducing new ML-based models aimed at CPR. The findings were classified into four categories according to CPR tasks: \textbf{(i)} Rhythm Analysis; \textbf{(ii)} Outcome Prediction; \textbf{(iii)} Non-Invasive Blood Pressure and Chest Compression; and \textbf{(iv)} Pulse and ROSC. In Figure \ref{fig:taxonomy}, we introduce a taxonomy for ML approaches used in CPR tasks, where we focus on the top four tasks and categorized the models into two main sections: ML and Deep Learning. In the ML category, we included basic and classic models, while in the deep learning category, we included models that use neural networks. We further divided the models into two types: standalone models and hybrid models. Standalone models use the core architecture without any significant changes, while hybrid models modify the core architecture. We identified several approaches that used standalone models for CPR tasks, as well as approaches that used hybrid models in both ML and deep learning.

The introduced taxonomy visualizes and provides a quick grasp of the existing ML approaches for different CPR tasks, research gaps, and possible ML approaches that can be experimented with for these tasks. We analyze all these approaches in detail and discuss additional ML techniques for CPR tasks in this section.

\subsection{Rhythm Analysis}

Accurate rhythm analysis during CPR is crucial for determining the appropriate interventions. In fact, survival during OHCA relies on accurate shock decisions from AEDs. However, chest compressions corrupt electrocardiogram (ECG) signals, complicating shock timing. Thus, ML methods are introduced to ensure the balance between continuous compressions and timely shock delivery.

Table \ref{tab:rhythm_analysis} depicts the existing ML approaches for rhythm analysis. \cite{DBLP:journals/isci/GongWYZZL23} propose a solution that integrates the UNet model for the restoration of corrupted ECG signals and ResNet model for the classification of shock/no-shock cases during chest compression. The model is assessed on datasets of different sizes, with transfer learning applied to the UNet model to mitigate overfitting. Similarly,~\cite{DBLP:journals/entropy/IsasiIAEKW20} propose a solution that assists in making accurate shock or no-shock decisions despite CPR-related imperfections, identically to the study conducted by  \cite{jaureguibeitia2020shock}. Within the same context, \cite{isasi2021shock} use SVM for binary shock decision-making after removing load distributing band (LDB) artifacts from ECG signals. In contrast, \cite{hajeb2022enhancing, hajeb2021deep} utilize a CNNs-based encoder-decoder approach to eliminate artifacts introduced during chest compression, aiming for precise shock decisions. In their study, \cite{jekova2021optimization} examine multiple CNN models to identify the most effective one for accurately determining whether to provide a shock or not when dealing with corrupted ECG signals.

From a different perspective, 
 \cite{picon2019mixed} suggest an approach that merges CNN and LSTM to identify lethal ventricular arrhythmia. The detection of this abnormal rhythm is crucial since it instructs the AED to apply shocks to return the heart to a normal rhythm or to continue performing CPR until normal cardiac rhythm has been restored.
In the same context, \cite{rad2017ecg} evaluate various ML approaches and found that Artificial Neural Networks (ANN) is the most effective for classifying the five resuscitation cardiac rhythms : Ventricular Tachycardia (VT), Ventricular Fibrillation (VF), Pulseless Electrical
Activity (PEA), Asystole (AS), and Pulse-generating Rhythm (PR). This classification aids in determining whether to deliver a shock during resuscitation based on the rhythm category.

\setlength\LTleft{-1.7cm}
\begin{longtable}{|>{\centering\arraybackslash}p{2cm}|>{\centering\arraybackslash}p{3cm}|>{\centering\arraybackslash}p{2cm}|>{\centering\arraybackslash}p{3cm}|>{\centering\arraybackslash}p{4.7cm}|}
\caption{Comparative Overview of ML Approaches for Rhythm Analysis Tasks of CPR.} \label{tab:rhythm_analysis} \\
\hline
\textbf{Reference} & \textbf{AI/ML Techniques} & \textbf{ML Class} & \textbf{Key Contribution} & \textbf{Task Description} \\
\hline
\endfirsthead

\multicolumn{5}{c}%
{{\bfseries \tablename\ \thetable{} -- continued from previous page}} \\
\hline
\textbf{Reference} & \textbf{AI/ML Techniques} & \textbf{ML Class} &  \textbf{Key Contribution} & \textbf{Task Description} \\
\hline
\endhead

\hline \multicolumn{5}{|r|}{{Continued on next page}} \\
\hline
\endfoot

\hline
\endlastfoot

\citep{DBLP:journals/isci/GongWYZZL23} & UNet, ResNet, Transfer Learning &  Self-supervised/ Supervised Learning & Signal restoration \& Accurate shock advice algorithm & Restoring corrupted ECG signals and aiding in shock/no-shock decision\\
\hline
\citep{hajeb2021deep} & CNN, ResNet, BiLSTM & Supervised Learning   & Accurate shock advice algorithm & Helping to accurately decide on delivering shocks or not during CPR\\
\hline
\citep{DBLP:journals/entropy/IsasiIAEKW20,jaureguibeitia2020shock} & CNN & Supervised Learning & Accurate shock advice algorithm & Aiding in making accurate shock or no-shock decisions during CPR despite CPR-related artifacts \\
\hline
\citep{rad2017ecg} & Bayesian Decision Theory, K-NN, K-local Hyperplane Distance Nearest Neighbor, ANN & Supervised Learning &  Improving the monitoring of heart rhythm during resuscitation & Assisting in the  classification of resuscitation cardiac rhythms using ECG data \\
\hline
\cite{picon2019mixed} & CNN-LSTM &  Supervised Learning  &  Detection of shockable arrhythmia & Aiding in the recognition of lethal ventricular arrhythmia during cardiac arrest through the analysis of ECG data\\
\hline

\citep{jekova2021optimization} & CNN & Supervised Learning  & Accurate shock advice algorithm  & Exploring the most effective CNN model for precise shock advisory decisions \\
\hline
\citep{isasi2021shock} & SVM & Supervised Learning  & Accurate shock advice algorithm & Filtering LDB artefacts and aiding shock decision-making during CPR \\
\hline
\citep{hajeb2022enhancing} & CNN-based Encoder-Decoder & Supervised Learning  & Accurate shock advice algorithm & Removing CPR-related artifacts for better shock decision-making \\
\hline
\end{longtable}

\textbf{Dataset:} Regarding the used data, the cited works are divided into two categories. Some of the papers \citep{DBLP:journals/entropy/IsasiIAEKW20,jaureguibeitia2020shock,rad2017ecg,
jekova2021optimization,isasi2021shock} used proprietary datasets that can be accessible but upon request to the owner, while the other category of papers \citep{DBLP:journals/isci/GongWYZZL23,picon2019mixed,hajeb2021deep,hajeb2022enhancing} used publicly available datasets. In most cases, the publicly used datasets are the Creighton University Tachyarrhythmia Database (CUDB) \citep{nolle1986crei}, the Massachusetts Institute of Technology‐Beth Israel Hospital Malignant Ventricular Arrhythmia Database (VFDB) \citep{greenwald1986development}, the Sudden Cardiac Death Holter Database (SDDB) \citep{greenwald1986development}, and the MIT-BIH Atrial Fibrillation Database (AFDB) \citep{goldberger2000physiobank}.
\\
\subsection{Outcome Prediction}

Outcome prediction in CPR involves analyzing the ECG signal characteristics, particularly before and after delivering CPR, to estimate the probability of successful defibrillation, survival, and neurological outcomes. Measures like the cardioversion outcome prediction (COP) quantify the likelihood of successful defibrillation by assessing ECG signal changes induced by CPR. Clinical prediction models using factors like initial presentation, interventions, and time intervals can also reliably estimate survival probabilities in OHCA patients \citep{survey01}. Accurate outcome prediction can guide appropriate interventions and management decisions during resuscitation efforts.\\
Table \ref{tab:outcome_prediction} analyzes the research paper while different ML techniques are being applied for outcome prediction tasks in CPR.  \cite{chen2022predicting} proposed a method to predict invasive coronary perfusion pressure using noninvasive ECG and photoplethysmography based on ML.  \cite{harford2019machine} designed a ML-centered model to forecast survival after cardiac arrest by sorting OHCA patients into two groups: those experiencing positive neurological outcomes and those encountering adverse neurological outcomes or mortality. \cite{howe2014support} utilized VF waveform metrics and SVM for predicting defibrillation success, enhancing CPR by optimizing the timing of defibrillation. Entropy-based features were employed by another study \citep{chicote2016application} to predict defibrillation success in cardiac arrest, improving shock outcome prediction accuracy. \cite{he2016combining} combined amplitude spectrum area with previous shock information using neural networks, enhancing the predictability of defibrillation outcomes in OHCA scenarios. In a similar advancement, \cite{shandilya2016integration} developed an MDI model that incorporates non-linear dynamics for robust prediction of defibrillation success, outperforming traditional methods. In this context, the work of  \cite{coult2019ventricular} represents a significant advancement. They developed an algorithm capable of predicting defibrillation outcomes during chest compressions, achieving an Area Under the Receiver Operating Characteristic (AUROC) of $0.74$ with chest compression artifacts, compared to $0.77$ without. This high level of calibration suggests that the algorithm could serve as a reliable probability index for defibrillation outcome predictions. As of now, no studies have been published on the use of deep learning for this purpose, but the recent work by \cite{ivanovic2020predicting} using CNN for ECG signal analysis indicates promising directions. Additionally, the integration of logistic regression and SVM by \cite{coult2021method} to analyze ECG data during CPR without needing to stop compressions, combining predictive modeling for both short-term and long-term outcomes, showcases the continuous evolution of ML applications in improving CPR effectiveness.
\setlength\LTleft{-1.7cm}
\begin{longtable}{|>{\centering\arraybackslash}p{2.2cm}|>{\centering\arraybackslash}p{3.5cm}|>{\centering\arraybackslash}p{2cm}|>{\centering\arraybackslash}p{3cm}|>{\centering\arraybackslash}p{4.5cm}|> {\centering\arraybackslash}p{1.5cm}|}
\caption{Comparative Overview of ML approaches for Outcome Prediction Tasks of CPR.} \label{tab:outcome_prediction} \\
\hline
\textbf{Reference} & \textbf{AI/ML Techniques} & \textbf{ML Class} & \textbf{Key Contribution} & \textbf{Task Description} \\
\hline
\endfirsthead

\multicolumn{6}{c}%
{{\bfseries \tablename\ \thetable{} -- continued from previous page}} \\
\hline
\textbf{Reference} & \textbf{AI/ML Techniques} & \textbf{ML Class} &  \textbf{Key Contribution} & \textbf{Task Description} \\
\hline
\endhead

\hline \multicolumn{6}{|r|}{{Continued on next page}} \\
\hline
\endfoot

\hline
\endlastfoot

\citep{coult2019ventricular} & SVM & Supervised Learning & Evaluated the predictive performance of VF waveform measures during ongoing chest compressions &  Combine multiple VF waveform measures, providing dynamic prognostic assessment.\\
\hline
\citep{harford2019machine} & Embedded Fully Convolutional Network (EFCN) &  Supervised Learning &  Prediction of neurological outcomes of patients & 
Classifying cardiac arrest patients into two groups based on their neurological condition \\
\hline
\citep{chen2022predicting} & Support vector regression (SVR), Random forest (RF), K nearest neighbor (KNN), Gradient boosted regression tree (GBRT) &  Supervised Learning & Prediction of invasive coronary perfusion pressure (CPP) & Prediction of coronary perfusion pressure (CPP) during CPR to guide chest compression process \\
\hline
\citep{shandilya2016integration} & Multiple Domain Integrative (MDI) Model, Non-linear Dynamics & Supervised Learning & Developed an MDI model for robust prediction of defibrillation success. & Enhance the accuracy and reliability of predicting defibrillation outcomes.\\
\hline
\citep{he2016combining} & Neural Networks & Supervised Learning & Improved prediction performance of defibrillation outcome. & Combine amplitude spectrum area with previous shock information, enhancing the predictability of defibrillation outcomes. \\
\hline

\citep{chicote2016application} & Entropy-based Features (Fuzzy Entropy) &Supervised Learning & Predict defibrillation success in cardiac arrest, improving shock outcome prediction accuracy. &  Analyze ECG data for predicting defibrillation success, demonstrating enhanced sensitivity and specificity in shock outcome prediction. \\
\hline
\citep{howe2014support} & SVM & Supervised Learning & Utilized VF waveform metrics and predicting defibrillation success, enhancing CPR by optimizing the timing of defibrillation. & Prediction of defibrillation success in VF cardiac arrest patients to process waveform metrics, achieving prediction accuracy \& generalization performance.\\
\hline
\citep{ivanovic2020predicting} & CNN & Supervised Learning  & Predicting defibrillation success by analyzing ECG signals, achieving high accuracy and implementation. & Automatically learn predictive features from ECG signals, crucial for guiding timely and effective CPR interventions. \\
\hline
\citep{he2016combining} & Neural Networks & Supervised Learning & Improved prediction performance of defibrillation outcome. & Combine amplitude spectrum area with previous shock information, enhancing the predictability of defibrillation outcomes. \\
\hline
\citep{coult2021method} & LR/SVM & Supervised Learning & Developed an algorithm to predict outcomes of VF shock during CPR, enhancing decision-making in real-time. & Uses ECG data analysis during CPR without needing to stop compressions, predict successful defibrillation and patient survival.\\
\hline

\end{longtable}
\textbf{Dataset:}
Most of the cited papers used private datasets. Exceptionally, \cite{he2016combining,shandilya2016integration,ivanovic2020predicting} provide samples of their used datasets for training and testing their models. \cite{he2016combining} provide a spreadsheet file containing data collected from the emergency departments of Southwest Hospital and Xinqiao Hospital in Chongqing between January $2012$ and February $2014$. \cite{shandilya2016integration} used data provided by the Richmond Ambulance Authority (Richmond, VA) and Zoll Medical Corp. (Chelmsford, MA), while \cite{ivanovic2020predicting} used publicly available ECG data.
\\

\subsection{Non-Invasive Blood Pressure and Chest Compression}

Non-Invasive blood pressure is a method to measure blood pressure without needing to insert anything into the body. It uses external devices like an inflatable cuff around the arm. During CPR, monitoring non-invasive blood pressure helps track the patient’s blood pressure to make sure chest compressions and other treatments are working \citep{Non_invensive}. Chest compressions are a key part of CPR. They involve pressing down on the patient’s chest to manually pump blood through the heart and circulate it around the body. Proper chest compressions are crucial to keep blood flowing to vital organs until the heart can start beating effectively on its own again \citep{chest_compress}. So, The non-invasive blood pressure and chest compression are co-related and crucial tasks of CPR. 
Table \ref{tab:MachineDrivenInnovations3} lists relevant ML approaches for non-invasive blood pressure and chest compression tasks. \cite{zhao2021recognition} utilize a CNN to examine chest compression depth data, distinguishing between abnormal and normal compressions to ensure CPR effectiveness. Meanwhile, \cite{jalali2014modeling} employ unsupervised ML to identify the optimal parameters describing chest mechanical properties during CPR. This effort aims to comprehend the intricate nature of CPR, ultimately enhancing its performance. On the other hand, \cite{lampe2019towards} harness the power of random forest algorithms to develop a predictive model that estimates carotid blood flow from chest compression parameters. Their work not only predicts but validates these estimations, paving the way for more tailored CPR techniques. Additionally, \cite{sebastian2020closed} apply linear regression within a closed-loop machine-controlled CPR system that dynamically adjusts compression characteristics based on real-time hemodynamic feedback. This innovative approach optimizes coronary perfusion pressure during prolonged CPR sessions, showcasing the potential of ML to improve long-term CPR effectiveness.

Similarly, a diverse array of algorithms including LSTM, neural networks, linear regression, and sparse spectrum gaussian process are used by \cite{blood_pressure_1} to demonstrate state-of-the-art modeling performance in predicting coronary perfusion pressure during CPR. This study analyzes the performance of each algorithm for single-step and long-term predictions, contributing significantly to our understanding of dynamic physiological responses during cardiac emergencies.

Additionally, another study by \cite{blood_pressure_2} has designed a system that utilizes LSTM for carrying out earlobe photoplethysmography, aiming at the non-invasive estimation of blood pressure. This innovative approach holds potential for enhancing patient monitoring in real-time during CPR. 

\setlength\LTleft{-1.7cm}
\begin{longtable}{|>{\centering\arraybackslash}p{2.2cm}|>{\centering\arraybackslash}p{4cm}|>{\centering\arraybackslash}p{2cm}|>{\centering\arraybackslash}p{3cm}|>{\centering\arraybackslash}p{4cm}|> {\centering\arraybackslash}p{1.5cm}|}

\caption{Comparative Overview of ML approaches for Non-Invensive Blood Pressure and Chest Compression Tasks of CPR.} \label{tab:MachineDrivenInnovations3} \\
\hline
\textbf{Reference} & \textbf{AI/ML Techniques} & \textbf{ML Class} & \textbf{Key Contribution} & \textbf{Task Description} \\
\hline
\endfirsthead

\multicolumn{6}{c}%
{{\bfseries \tablename\ \thetable{} -- continued from previous page}} \\
\hline
\textbf{Reference} & \textbf{AI/ML Techniques} & \textbf{ML Class} &  \textbf{Key Contribution} & \textbf{Task Description} \\
\hline
\endhead

\hline \multicolumn{6}{|r|}{{Continued on next page}} \\
\hline
\endfoot

\hline
\endlastfoot

\citep{blood_pressure_1} & LSTM, Neural Network, Linear Regression, Sparse Spectrum Gaussian Process & Supervised Learning & Analyze the performance of each algorithm for single-step and long-term predictions & Predict coronary perfusion pressure during CPR. \\
\hline

\citep{jalali2014modeling} & K-NN &  Supervised Learning &  Modeling mechanical properties of the chest & Finding the best set of parameters which describe the chest mechanical properties during CPR \\
\hline

\citep{sebastian2020closed} & Linear Regression & Supervised Learning &  Optimizing coronary perfusion pressure during CPR  & Implementing machine-controlled system to adjust compression characteristics. \\
\hline

\citep{lampe2019towards} & Random Forest & Supervised Learning &  Developed a model to predict carotid blood flow from chest compression parameters & Validation of a predictive model to estimate carotid blood flow, the duration of resuscitation.\\
\hline

\citep{blood_pressure_2} & LSTM & Supervised Learning & Designed a system for estimation of blood pressure & Estimation of arterial blood pressure. \\

\hline

\citep{zhao2021recognition} & CNN & Supervised Learning &  Recognizing abnormal chest compression depth & Analyzing chest compression depth data to ensure effective CPR \\
\hline

\end{longtable}

\textbf{Dataset:}
All the cited work used proprietary datasets. Some datasets contain human cases of cardiac arrest victims \citep{sebastian2020closed} provided by medical institutions. In \citep{zhao2021recognition}, they use datasets collected internally, while in \citep{lampe2019towards, jalali2014modeling, blood_pressure_2, blood_pressure_1}, animal data such as porcine data were used for their studies.

\subsection{Pulse and Return Of Spontaneous Circulation Detection (ROSC)}

In CPR, pulse detection is an important aspect that assists in deciding whether to continue chest compressions or to pursue alternative interventions. In the context of CPR, pulse detection and ROSC detection are closely related as they both involve assessing the presence or absence of a pulse or heartbeat to guide resuscitation efforts \citep{pulse_00}. Once chest compressions are initiated, the healthcare team continues to monitor the patient for signs of ROSC. ROSC occurs when a patient who previously had no pulse or heartbeat regains a spontaneous, organized cardiac rhythm. Detecting ROSC during CPR is a critical milestone as it indicates that circulation has been restored and that the heart is functioning on its own \citep{ROSC}.

In the context of CPR, ROSC detection is closely related as they both involve assessing the presence or absence of a pulse or heartbeat to guide resuscitation efforts, Table \ref{tab:MachineDrivenInnovations4} noted the contributing ML techniques for this task execution. \cite{isasi2021machine} and \cite{sashidhar2021machine}  utilize ML techniques to identify a pulse seamlessly during CPR, aiming to avoid disruptions in the process. Their approaches aim to ensure continuous and effective resuscitation efforts. In pursuit of minimizing interruptions in the CPR process, \cite{DBLP:journals/access/AlonsoIAD20} introduce a reliable classifier for prompt pulse/no-pulse decisions, with the potential to enhance survival rates. Recently, there has been a notable introduction of the concept of pseudo-PEA, characterized by organized electrical activity accompanied by mechanical activity that is insufficient to produce a detectable pulse \citep{ROSC_2}. \cite{ROSC_3} conducted a subsequent investigation, developing a Random Forest algorithm to distinguish between ROSC, Pulseless Electrical Activity (PEA), and pseudo-PEA. Additionally, these advancements highlight the potential of ML in analyzing intricate signals and extracting crucial information, thereby improving decision-making in emergency cardiac care.

In another context, \citep{shao2022intelligent} is among the scarce studies employing Reinforcement Learning (RL) in CPR. They introduce a deep RL model grounded on a Deep Q-Network (DQN) policy to intelligently diagnose and recommend treatment for cardiac arrest instances, even in scenarios of incomplete or missing patient information. Their model aims to optimize CPR success rates and sustain ideal blood pressure levels during pulse detection and ROSC.

\setlength\LTleft{-1.7cm}
\begin{longtable}{|>{\centering\arraybackslash}p{2cm}|>{\centering\arraybackslash}p{4cm}|>{\centering\arraybackslash}p{2.45cm}|>{\centering\arraybackslash}p{3cm}|>{\centering\arraybackslash}p{3.5cm}|> {\centering\arraybackslash}p{2cm}|}

\caption{Comparative Overview of ML approaches for Pulse and ROSC Tasks of CPR.} \label{tab:MachineDrivenInnovations4} \\
\hline
\textbf{Reference} & \textbf{AI/ML Techniques} & \textbf{ML Class} & \textbf{Key Contribution} & \textbf{Task Description} \\
\hline
\endfirsthead

\multicolumn{6}{c}%
{{\bfseries \tablename\ \thetable{} -- continued from previous page}} \\
\hline
\textbf{Reference} & \textbf{AI/ML Techniques} & \textbf{ML Class} & \textbf{Key Contribution} & \textbf{Task Description} \\
\hline
\endhead

\hline \multicolumn{6}{|r|}{{Continued on next page}} \\
\hline
\endfoot

\endlastfoot

\citep{DBLP:journals/access/AlonsoIAD20} & SVM &  Supervised Learning & Pulse detection & Accurately detecting pulses using defibrillator signals \\
\hline

\citep{ROSC_4} & Linear discriminant analysis, Quadratic discriminant analysis, SVM, Neural networks, Logistic regression, Random Forest & Supervised Learning & Predicting pulse status & Using sensor data to predict the presence or absence of a pulse during chest compressions  \\
\hline

\citep{ROSC_2} & Random Forest & Supervised Learning & Pulse detection & Detection of return of spontaneous circulation in OHCA \\
\hline

\citep{isasi2021machine} & Random Forest & Supervised Learning & Pulse detection & Detection of pulse using the ECG and Thoracic Impedance (TI) signals without interrupting CPR process \\
\hline

\citep{shao2022intelligent} & DQN &  Reinforcement Learning & Improving cardiac emergency care & Assisting in treating cardiac arrest with incomplete patient information.\\
\hline
\end{longtable}

\textbf{Dataset:}
Most of the data for pulse detection and ROSC (Return of Spontaneous Circulation) tasks were gathered from hospital patients over an extended period, with the support of various governing and medical authorities. Consequently, much of the experimental datasets remain confidential. However, access to these datasets can be arranged by contacting the authors of the relevant research papers as indicated in \citep{isasi2021machine} \citep{DBLP:journals/access/AlonsoIAD20} \citep{ROSC_2} \citep{ROSC_3} \citep{shao2022intelligent}. However, a set of public data is available for pulse prediction tasks, which evaluated 383 patients being treated for OHCA \citep{ROSC_4}. The data repository can be accessed at \footnote{\href {https://doi.org/10.5281/zenodo.3995071}{www.zenodo.org}}.
\\

Table \ref{tab:MachineDrivenInnovations5} shows and explores several other ML approaches for other CPR tasks. For instance, \cite{blomberg2019machine}, \cite{blomberg2021effect},  \cite{chin2021early}, and \cite{byrsell2021machine} investigated the use of ML as a supportive tool to recognize cardiac arrest in emergency calls, highlighting the potential of AI in early intervention using supervised method. \cite{bender2020machine} proposed a ML algorithm to improve patient-centric pediatric CPR.

\setlength\LTleft{-1.7cm}
\begin{longtable}{|>{\centering\arraybackslash}p{2.2cm}|>{\centering\arraybackslash}p{4cm}|>{\centering\arraybackslash}p{2cm}|>{\centering\arraybackslash}p{3cm}|>{\centering\arraybackslash}p{4cm}|>{\centering\arraybackslash}p{1.5cm}|}
\caption{Comparative Overview of ML Approaches for Other Tasks of CPR.} \label{tab:MachineDrivenInnovations5} \\
\hline
\textbf{Reference} & \textbf{AI/ML Techniques} & \textbf{ML Class} & \textbf{Key Contribution} & \textbf{Task Description} \\
\hline
\endfirsthead

\multicolumn{6}{c}%
{{\bfseries \tablename\ \thetable{} -- continued from previous page}} \\
\hline
\textbf{Reference} & \textbf{AI/ML Techniques} & \textbf{ML Class} & \textbf{Key Contribution} & \textbf{Task Description} \\
\hline
\endhead

\hline \multicolumn{6}{|r|}{{Continued on next page}} \\
\hline
\endfoot

\hline
\endlastfoot

\citep{bender2020machine} & SVM & Supervised Learning & Enhance child-focused CPR care & Identify pediatric ventricular fibrillation types. \\
\hline
\citep{blomberg2019machine} & 
Automatic speech recognition (ASR) model for speech-to-text and an OHCA detector & Supervised Learning & Early detection of cardiac arrest & Detection of cardiac arrest during emergency phone calls using audio analysis \\
\hline
\citep{blomberg2021effect} & ASR model for speech-to-text and an OHCA detector & Supervised Learning & Improved dispatcher recognition & Enhancing dispatcher's ability to recognize cardiac arrest during emergency calls\\
\hline
\citep{byrsell2021machine} & ASR model for speech-to-text and an OHCA detector for real-time prediction & Supervised Learning & Support dispatchers in cardiac arrest recognition & Assisting dispatchers in identifying cardiac arrest from emergency call data\\
\hline
\citep{chin2021early} &  SVM & Supervised Learning & Early recognition of caller’s emotion & Analyzing caller's emotions during emergency calls to aid in early cardiac arrest detection \\
\hline
\end{longtable}

\textbf{Datasets:}
In the case of the other tasks, we found that all of the experimental data were collected privately from cardiac patients at hospitals or out-of-hospital, with assistance from local medical or governing authorities like the Danish Cardiac Arrest Registry. Specifically, data to identify cardiac attacks through emergency calls were obtained using a modified dispatch data survey catalog. However, data for this CPR task are not publicly available and remain private considering patients' privacy.\\

The studies discussed in this section underscore the potential of ML and AI in revolutionizing CPR. From early detection to intervention and post-event analysis, these technologies pave the way for more personalized and effective resuscitation strategies.

\section{Future Prospects, Challenges and Discussion}
\label{section5}

After thoroughly exploring current ML methods for improving CPR, we have looked into the prospects of how these approaches can naturally make CPR more effective. We examined various existing ML strategies and identified those paying close attention to supervised learning. However, the availability of labeled data is very difficult considering real-life CPR scenarios. In contrast, there is limited experimentation with unsupervised and self-supervised learning approaches. In addition, only one paper has been identified concerning the application of the RL approach for CPR optimization. In fact, the potential for the utilization of RL in automating and optimizing CPR procedures is still not explored significantly. Moreover, current CPR research employs traditional ML methods like Random Forest, KNN, SVM, LSTM, and CNN, but lacks exploration of Transformer-based architectures, which might offer superior capabilities in handling sequential and contextual data, potentially revolutionizing resuscitation research. Furthermore, Human-aware AI and explainability are important concerns in the future, especially in considering critical decision-making tasks in medical applications in general \citep{ZakiQK23} and the CPR context using ML in particular.  In this section, we will investigate all of these possibilities that RL could offer for CPR, as well as explore other AI techniques such as transformers and Explainable AI (XAI), highlighting their significance in the context of CPR.

\subsection{Potentiality of Reinforcement Learning}

\begin{figure}[!htp]
    \begin{center}
    \makebox[\textwidth][l]{\hspace*{-1.3cm}\includegraphics[width=1.25\textwidth,height=.7\textwidth]{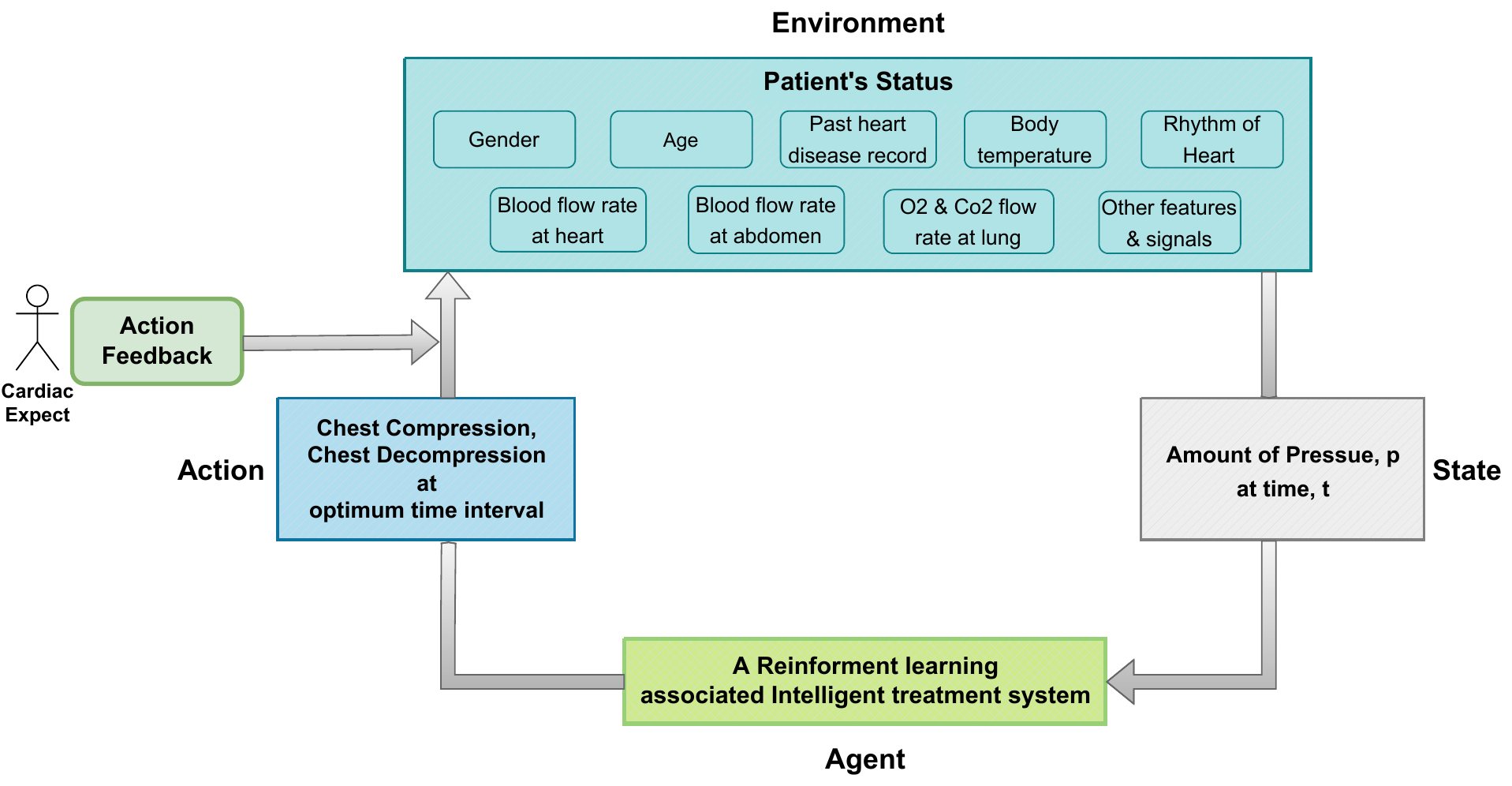}}
    \caption{A human feedback reinforcement learning associated CPR optimization system.}
    \label{fig:future_work}
    \vspace{-10pt} 
    \end{center}
\end{figure}

RL algorithms can be effective in learning optimal chest compression techniques by interacting with simulated or real-world CPR scenarios. The RL agent, representing the model, can receive feedback on compression depth, rate, and the patient's other status, dynamically adapting its actions to maximize the likelihood of successful resuscitation and provide the optimal amount of pressure on the chest and abdomen to increase the blood pressure on the heart, where the time interval is also very crucial and RL model identify the optimal pressure and time step, Figure \ref{fig:future_work}.  Human feedback reinforcement learning (HFRL) can be effective in CPR by incorporating real-time guidance from human practitioners. In CPR scenarios, HFRL can adaptively learn and refine CPR techniques based on direct feedback from healthcare providers, improving the accuracy and effectiveness of CPR administration over time. This approach leverages medical expertise to enhance the learning process of RL algorithms, ensuring that CPR techniques are continually optimized and tailored to individual patient needs and response patterns \citep{RFRL_00, RFRL_01}. Moreover, RL is instrumental in the development of adaptive CPR protocols that adjust dynamically based on patient responses and evolving clinical conditions. By recognizing patterns in physiological data, the RL agent can adjust compression parameters and recommend interventions accordingly. Analyzing large datasets with RL algorithms can help to identify patterns that link to successful outcomes, which aids in continuously improving quality, refining protocols, and updating guidelines based on real-world results  \citep{RL_00, RL_01}. Furthermore, RL algorithms have the capability to analyze patient-specific data and develop personalized resuscitation plans, leading to CPR strategies customized to individual patient characteristics. Multiple RL methods can effectively accomplish these tasks \citep{RL_02, RL_03}.

\textbf{Policy Gradient Methods:} Policy Gradient Methods play a crucial role in training adaptive policies that guide healthcare providers during resuscitation efforts. These methods learn from observed states, such as patient vital signs, to directly map to optimal actions, like adjusting chest compression depth and rate. They are well-suited for continuous and dynamic action spaces, making them effective for improving the quality of chest compressions \citep{future_policy_01}. By shaping reward functions to encourage desirable actions, addressing uncertainty, and accommodating patient-specific conditions, these methods contribute to personalized and effective CPR strategies. Their adaptability, online learning capabilities, and potential for transfer learning across scenarios make Policy Gradient Methods valuable tools for enhancing decision-making in dynamic and critical CPR environments \citep{future_policy_02}, \citep{future_policy_03}.

Moreover, Actor-critic methods combine aspects of both value-based (Q-learning) and policy-based methods. The actor (policy) learns the optimal actions, and the critic evaluates the chosen actions. This can be useful for learning adaptive resuscitation protocols \citep{future_actor_01}, \citep{future_Q_01}.

\textbf{Policy Optimization Algorithms and Deep Deterministic Policy Gradient (DDPG):} 
In the field of CPR optimization, Policy Optimization Algorithms and Deep Deterministic Policy Gradient (DDPG) are particularly effective algorithms for learning directly the best actions to take during resuscitation based on the observed states, which can be useful for dealing with the continuous and dynamic nature of CPR such as adjusting chest compression depth and rate \citep{future_POA_01}, \citep{future_POA_02}. Both Policy Optimization Algorithms and DDPG support adaptive and personalized CPR strategies by shaping reward functions, addressing uncertainties, and enabling continuous online learning \citep{future_DDPG_01}, \citep{future_DDPG_02}.

\textbf{Meta-reinforcement learning and Inverse RL:} 
Meta-reinforcement learning (Meta-RL) and Inverse Reinforcement Learning (IRL) offer innovative solutions to diverse problems. Meta-RL has the potential to adapt to new CPR scenarios quickly with minimal data, ensuring effective decision-making in various resuscitation situations \citep{future_metaRL_01}. Besides, IRL might be capable of addressing the challenge of learning optimal resuscitation strategies by observing expert demonstrations. It is potential to leverage these demonstrations to uncover underlying preferences or policies, providing valuable insights into effective CPR decision-making. Both of these approaches help develop models that can generalize effectively across different patient conditions \citep{future_InverseRL_01}, \citep{future_InverseRL_02}.

Furthermore, RL can help optimize the placement of automated external defibrillators (AEDs) in public spaces \citep{future_aed}. By learning from historical data on cardiac arrest incidents, RL algorithms can recommend the best locations for AED deployment, enhancing accessibility and reducing response times.\\

\subsection{Transformers and Explainable AI}
On the other hand, our investigation of ML approaches for future applications identified a significant gap. Although methods like Random Forest, KNN, SVM, LSTM, and CNN have been used in CPR research, experiments with Transformer-based architectures are notably absent. Transformers have shown promise in capturing and understanding sequential and contextual information in the long range, which can be effective in CPR scenarios. Their strong ability to analyze temporal and dynamic data suggests that incorporating Transformer-based architectures could greatly enhance the range of ML methods used in resuscitation research.\\

\textbf{Potentiality of Transformers:} Transformers, designed for the efficient processing of sequential data, hold significant promise in capturing long-range dependencies and temporal patterns crucial for decision-making, which can be vital in CPR signal processing \citep{future_transformer_01,LinWLQ22}. The attention mechanism might be useful to focus on specific parts of input CPR sequences, prioritizing relevant information in response to variations in patient responses during CPR interventions. Transformers are particularly skilled at capturing contextual information, which can be invaluable in comprehending physiological changes and the effectiveness of interventions in CPR scenarios. Moreover, with the ability to process data in parallel, Transformer models could provide quick and accurate real-time decision support, crucial in dynamic CPR situations \citep{future_transformer_02}. In addition, pre-training on large datasets enables transfer learning for CPR-specific tasks, leveraging knowledge from diverse healthcare contexts \citep{han2021pre,ruan2022survey}. The interpretability provided by attention weights promotes transparency, helping healthcare professionals comprehend the model's decision-making process. Additionally, Transformers can be useful to support continuous learning and adaptation to evolving patient conditions, updating their understanding during resuscitation efforts for ongoing decision support. They also exhibit flexibility in handling multi-modal inputs, integrating diverse data sources like ECG, blood pressure, and respiratory rate for comprehensive decision-making. Further exploration of Transformer model applications can potentially lead to groundbreaking improvements in CPR processing \citep{future_transformer_03,future_transformer_04}.\\

\textbf{Relevance of Explainable AI:} 
Looking forward into the future, at least two central themes for research linking ML to CPR seem to be of interest to make  progress. We have seen how ML algorithms may yield important guidance for administering CPR. But it will be practitioners who actually take actions and make decisions based on what the ML systems indicate. This means that providing effective interfaces for the human users to view what the ML systems propose will be quite important \citep{nuutinen2023systematic,bienefeld2023solving}. And as practitioners are increasingly asked to adjust their decisions according to the advice of AI systems, a really essential goal will be to support providing explanations of the reasoning of those AI systems (known as XAI or explainable AI) in a way that the human users can comprehend. The more that our AI systems migrate to embracing deep learning, the more challenging it will be to actually explain the inner workings of those systems \citep{contreras2023explanation}. If solutions based on Large Language Models (LLMs) really do become more commonplace for any healthcare applications of ML, there are at least some ideas currently for how to generate those explanations \citep{huang2023can} as a promising starting point.  The introduction of attention-based transformer models can highlight the relevance of particular contexts in use but here too some XAI research has explored how to work with these approaches \citep{chefer2021generic,alammar2021ecco} so that XAI within CPR  should be effective as well. The fact that other researchers have begun to tackle human-AI partnerships in the setting of emergency medicine is also very encouraging \citep{okada2023explainable,maatta2023diagnostic}. The study of human-aware AI may pose challenges to resolve in full and systems in use that yield more straightforward output are certainly important as our first steps, but the path ahead with more forward-looking AI models is one that we can seek to explore, in order to secure important acceptance of these systems in all their forms, into the future.

\subsection{Discussion and Conclusion}
In future CPR research, a major obstacle is the difficulty of obtaining real-life CPR data. Accessing high-quality, labeled data for CPR scenarios is challenging due to privacy regulations and ethical considerations surrounding medical data. In addition, it is difficult to get large amounts of data classifying CPR procedures into successful and unsuccessful administrations. This scarcity of large and diverse datasets that accurately represent various CPR situations makes it hard to train robust ML models. To overcome these challenges, generative models such as variational autoencoders can help create large amounts of data points resembling the existing small amount of real-life data. Moreover, synthetic data generation techniques can be delicate along with real-life data, whereas more mathematical models such as Babbs provide representations of CPR scenarios, allowing the generation of realistic synthetic data \citep{babbs01, babbs00}. Additionally, data augmentation techniques offer a promising strategy to enhance datasets, enabling extensive research in CPR. The scarcity of data is not the only challenge; selecting the right data is also complicated. The use of Self-Directed Machine Learning (SDML) is one of the ML solutions that may be adopted in the future in CPR applications, as it supports medical staff in selecting proper data and making the right decisions in a secure and robust manner \citep{zhu2022self}.\\

To conclude, this paper highlights the potential of integrating ML with life-saving CPR techniques during cardiac arrest. By thoroughly reviewing and analyzing existing ML applications in CPR, we have identified key research gaps and unexplored opportunities that could pave the way for significant advancements in the biomedical field. Our comprehensive synthesis underscores the importance of interdisciplinary collaboration between computer science and medical research and sets a strategic framework for future studies. We believe that the continued exploration and implementation of ML in CPR will improve resuscitation outcomes, enhance the efficiency and effectiveness of emergency medical interventions, and ultimately save more lives.

\bibliography{sn-bibliography}

\end{document}